\documentclass[conference]{IEEEtran}
\IEEEoverridecommandlockouts

\newcommand{\eg}{\textit{e}.\textit{g}.}
\usepackage[numbers,sort&compress]{natbib}
\usepackage{amsmath,amssymb,amsfonts}
\usepackage{algorithmic}
\usepackage{graphicx}
\usepackage{textcomp}
\usepackage{xcolor}
\usepackage{float}
\usepackage{placeins}
\usepackage{enumitem}
\usepackage{balance}

\usepackage{diagbox}
\def\BibTeX{{\rm B\kern-.05em{\sc i\kern-.025em b}\kern-.08em
	T\kern-.1667em\lower.7ex\hbox{E}\kern-.125emX}}

	\title{Improving Human Image Synthesis with Residual Fast Fourier Transformation and Wasserstein Distance\\}
	
\DeclareRobustCommand*{\IEEEauthorrefmark}[1]{
	\raisebox{0pt}[0pt][0pt]{\textsuperscript{\footnotesize\ensuremath{#1}}}}
\begin{document}

\author{\IEEEauthorblockN{Jianhan Wu\IEEEauthorrefmark{1,2}, Shijing Si\IEEEauthorrefmark{1*}, Jianzong Wang\IEEEauthorrefmark{1} , Jing Xiao\IEEEauthorrefmark{1}}
	\IEEEauthorblockA{\emph{\IEEEauthorrefmark{1}Ping An Technology (Shenzhen) Co., Ltd.,} Shenzhen, China\\
		\IEEEauthorblockA{\emph{\IEEEauthorrefmark{2}University of Science and Technology of China,} Hefei, China  }
		Emails: wujianhan@mail.ustc.edu.cn, shijing.si@outlook.com, jzwang@188.com, xiaojing661@pingan.com.cn}
	\thanks{* Corresponding author: Shijing Si, shijing.si@outlook.com}
}
\maketitle

	\begin{abstract}
	With the rapid development of the Metaverse, virtual humans have emerged, and human image synthesis and editing techniques, such as pose transfer, have recently become popular. Most of the existing techniques rely on GANs, which can generate good human images even with large variants and occlusions. But from our best knowledge, the existing state-of-the-art method still has the following problems: the first is that the rendering effect of the synthetic image is not realistic, such as poor rendering of some regions. And the second is that the training of GAN is unstable and slow to converge, such as model collapse. Based on the above two problems, we propose several methods to solve them. To improve the rendering effect, we use the Residual Fast Fourier Transform Block to replace the traditional Residual Block. Then, spectral normalization and Wasserstein distance are used to improve the speed and stability of GAN training. Experiments demonstrate that the methods we offer are effective at solving the problems listed above, and we get state-of-the-art scores in LPIPS and PSNR.
	\end{abstract}
	
	
	\section{Introduction}
	Metaverse \cite{duan2021metaverse} has received comprehensive attention from around the world with the development of virtual reality (VR) and augmented reality (AR) \cite{xiong2021augmented}, which has eight attributions: immersion, low latency, Identity, friends, diversification, anytime and anywhere, economic system, civilization. Among them, immersion is the biggest attribution, which along with the technique of realistic and real-time full-body human image synthesis. So motivated, our goal is to synthesize a vivid and clear human image, and its posture can be arbitrarily changed.

	At present, Generative Adversarial Networks (GAN) \cite{goodfellow2014generative, si2021speech2video} has made great progress in realistic image synthesis. Some methods \cite{tang2020xinggan, li2020pona} are by extracting the features of different parts of the human body and then mapping from the original image pose to the target image pose. However, it is very difficult to generate a large-scale pose-transferred image by using the body region texture mapping as a representation, because sharp pose changes will cause the body to be non-corresponding with the texture. Creating invisible elements of the human body also poses a number of difficulties. In order to deal with these problems, some methods introduce a parsing map \cite{men2020controllable, zhang2021pise} to solve the problem of uncorresponding images caused by sharp poses. The advantage of the human parsing map can provide the semantic relevance of different parts to alleviate the problem of unaligned images. The disadvantage is that it cannot contain the shape and style features of the characters, and it does not contain the spatial characteristics of images, which results in the difficulty to generate realistic human images. To our best knowledge, the state-of-the-art method that can generate realistic human images and change their poses is PISE \cite{zhang2021pise}. PISE decouples the style of clothes and poses of humans and then trains them in two stages. The first stage is generating the human parsing map of the target pose through a parsing generator, the input is the keypoints of the source image (generated by openpose \cite{cao2019openpose}) and the semantic map (generated by PGN \cite{gong2018instance}) of the source image and the keypoints of the target image. The second stage is to fit the human parsing map above and the texture of the source image to generate the target image. In fact, the effect of the generated image from the method is always unsatisfactory, and the model is difficult to train.
	
	The aforementioned methods will encounter the following three challenges in generating satisfactory human images: (1) the generated image is satisfactory; (2) the generated effect is unstable, and GAN is difficult to train; (3) it cannot ideally change the pose of the human image.
	\begin{figure*}
		\centering
		\includegraphics[width=1.0\linewidth]{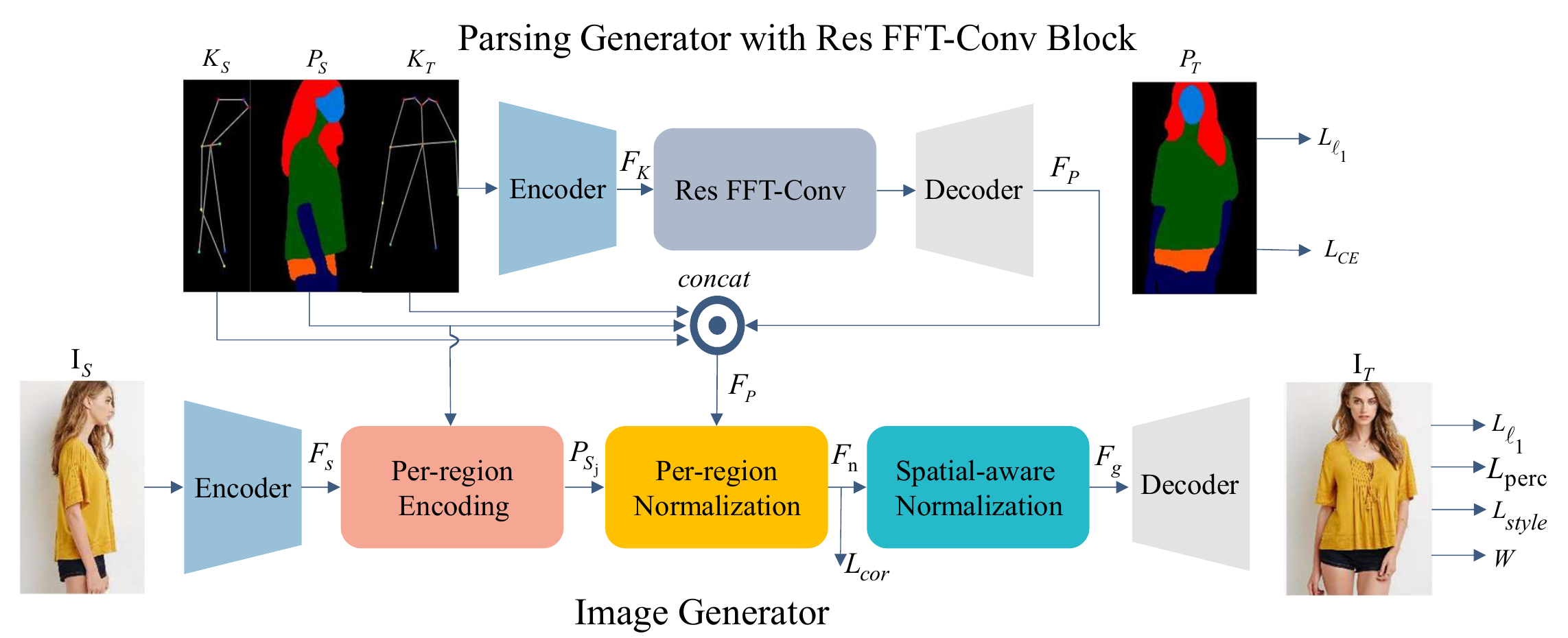}
		\caption{\emph{Overview of our model.} Our model consists of parsing generator and image generator, training parsing generator requires a pair of source-target images $ I_S, I_T$, then obtaining human keypoints $ K_S, K_T $ and human parsing map $ P_S, P_T$ respectively by using openpose and PGN framework. We concatenate $ K_S, P_S, K_T $ as the input of parsing generator, then the input is fed into an Unet-type network that generates a target parsing map with the same size of $ P_T$, which contains body shape information. To get the vivid image with detailed texture (\eg style of clothing), we extract the Per-region feature information $ P_{S_{j}}$ from the source image via VGG type network, then we concatenate the parsing map to the above $ K_S, P_S, K_T $ and normalize it along with the region information $ P_{S_{j}}$  to combine the information of source texture and target parsing map. Finally, the target image can be generated by spatial normalization and decoder.}
		\label{fig:fig}
	\end{figure*}
	
	 To solve these problems, we proposed a new structure based on the state-of-the-art method PISE. First of all, we analyse and discover the lack of adequate detailed information on existing methods through experiments. Then, we introduced the Residual Fast Fourier Transform with Convolution (Res FFT-Conv) \cite{mao2021deep} Block to replace the traditional ResBlock \cite{zhang2021drnet}. Although the traditional ResBlock may have a good ability in capturing high-frequency components of the image, it tends to ignore low-frequency information. The residual Fourier transform can capture long-term and short-term interactions while integrating spatial and frequency residual information, which is conducive to generating clear images. Finally, for the difficult training problem of GAN, we use Wasserstein distance \cite{arjovsky2017wasserstein} and spectral normalization \cite{miyato2018spectral} to solve it. Using Wasserstein distance can solve the problem of gradient disappearance, and using spectral normalization can limit the gradient of the discriminator not to exceed the Lipschitz constant k, which makes the discriminator satisfy Lipschitz continuity. The results of qualitative and quantitative experiments demonstrate that our method can not only guarantee the function of changing human pose arbitrarily but also improve its performance, reaching the best scores in LPIPS \cite{zhang2018unreasonable} and PSNR \cite{huynh2008scope}. Our contributions are summarized as follows:
	 \begin{itemize}[itemsep=8pt,topsep=8pt,parsep=0pt,partopsep=0pt,leftmargin=15pt]
	 	\item We use the Res FFT-Conv Block to replace the traditional Residual Block to improve the performance of generated images and analyze its applicability in principle. 
	 	\item Spectral normalization and Wasserstein distance are used to improve the speed and stability of GAN training. 
	 	\item We conduct abundant experiments to demonstrate the effectiveness of our method in comparison with some other baseline methods. And a comprehensive ablation study reveals the contribution of each part to the effectiveness improvement. 
	 \end{itemize}

	\section{Related Works}
	The pipeline we used involved lots of human image synthesis techniques. We are interested in the most effective jobs and applying them to our structure. 
	
	\subsection{Human Image Synthesis}
Human image synthesis technique has been widely studied, which is mainly divided into 3D images and 2D images. 3D human image synthesis mainly relies on 3D human models (such as SMPL \cite{loper2015smpl}). Human image synthesis and reconstruction are achieved by learning parameters of a 3D human model with deep neural networks \cite{zhang2021pymaf}. Two-dimensional image synthesis mainly relies on GAN. The generator and the discriminator are jointly restricted and upgraded under each iteration. A variety of images that are sufficiently clear and vivid can already be generated with the development of GAN. For example, pix2pix \cite{isola2017image} can convert one type of image to another type, which uses conditions such as semantic information to generate high-resolution images, and can edit the semantic information of the target images to achieve the function of editing images. PG$^2$ \cite{ma2017pose} first introduced pose transfer of human image, which concatenates the source image, source pose, and target pose as input to learn the target image, but the details of generated images by PG is not satisfactory. To alleviate the situation of generating image blur, they use a coarse-to-fine framework to deal with it. SCAGAN \cite{yu2021spatial} first alleviates the spatial misalignment by transferring the edge content to the target pose in advance, satisfactory human images are then synthesized by combining the appearance features of the source images, target pose keypoints, and prior content in the edge domain. The most recent developed work in human image synthesis is PISE. They decouple the shape of the character and the style of the clothes, and then use Per-region normalization \cite{zhu2020sean} and Spatial-aware normalization to achieve the functions of human pose transfer and texture transfer. However, the effect of the human image produced by this method is not ideal. In this paper, we found its problems through experiments and solved them by applying Res FFT-Conv Block and Wasserstein distance. 

	\subsection{Residual Fourier Transformation}
	Since Kaiming He proposed ResNet in image classification, the idea of introducing residuals has become very popular. The introduction of residual block can make the network deeper, have better generalization performance, and can effectively avoid gradient collapse \cite{chen2021exploring}. In the field of image synthesis, the residual idea is often transferred to use ResBlock, which can learn the difference between clear and blurred images. STDGAN \cite{zhang2019stdgan} uses ResBlock to denoise images. DRB-GAN \cite{xu2021drb} improves the effect of style transfer by using dynamic ResBlock. L2M-GAN \cite{yang2021l2m} uses ResBlock to learn to manipulate latent space semantics for facial attribute editing. These all indicate the excellent learning ability of ResBlock. However, ResBlock often only focuses on high-frequency information \cite{mao2021deep}, and generating clear photos requires both high-frequency information and low-frequency information. A lot of ResBlocks are used in the PISE network architecture, which leads to the unsatisfactory effect of the generated images. The Res FFT-Conv Block has an excellent performance in tasks such as denoising and image generation, and it can capture both high-frequency information and low-frequency information. Therefore, we use Res FFT-Conv Block instead of ResBlock to focus on the interaction of long-term and short-term information. The experiment results show that although the Res FFT-Conv Block is a simple plug-and-play module, it has a significant effect on improving image clarity and realism.
	
	\section{Methodology}
	We achieve the purpose of decoupling human pose and clothing style by using two generators, which are the parsing map generator with Res FFT-Conv Block and the image generator, and their parameters are updated and generated by different loss constraints. The overall framework is shown in Figure \ref{fig:fig}. Next, we introduce the generator, discriminator, and loss function one by one.

	\subsection{Parsing Generator With Res FFT-Conv Block}

\begin{figure}
		\centering
		\includegraphics[width=0.6\linewidth]{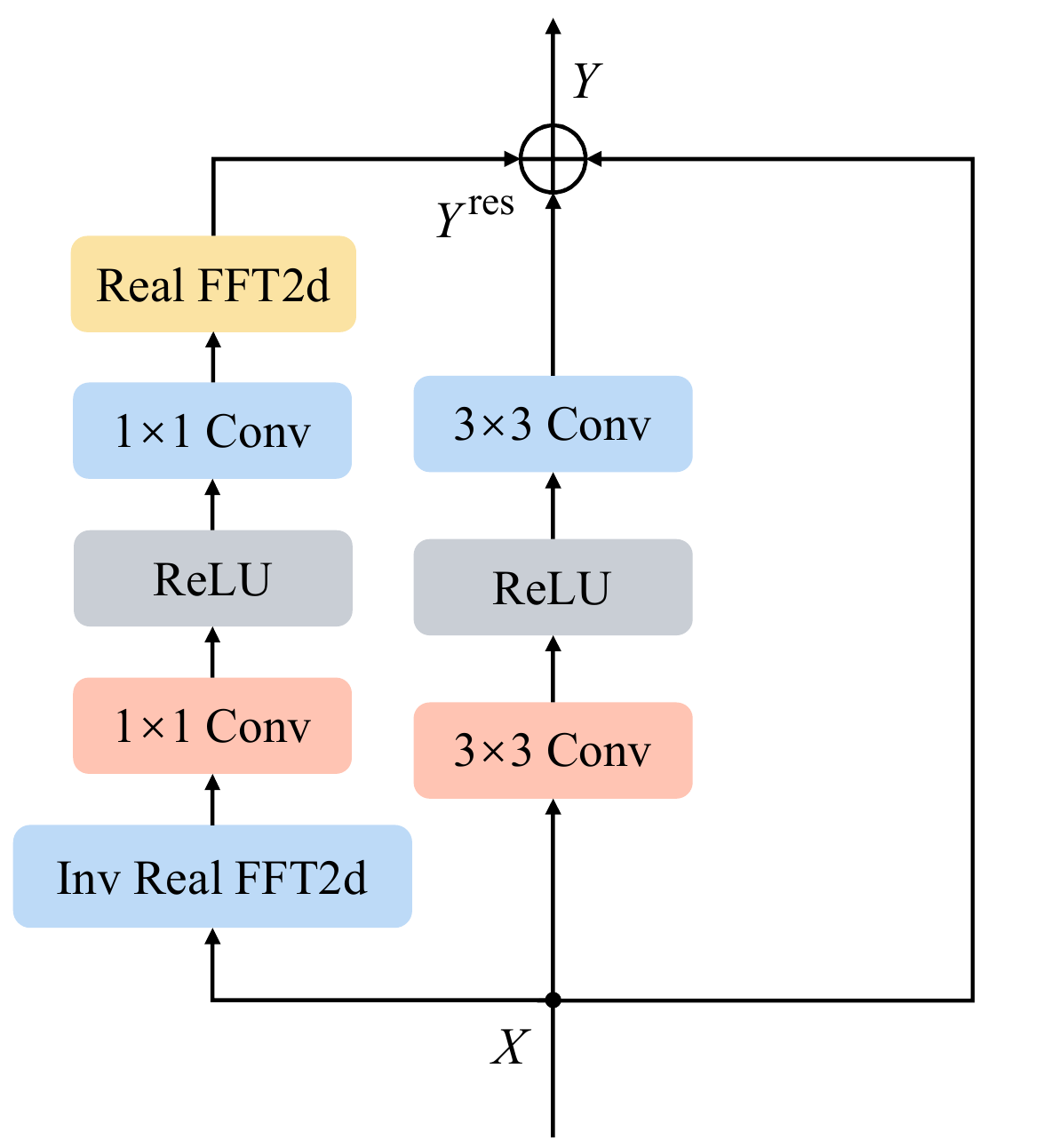}	
		\caption{Res FFT-Conv Block.}
		\label{fig4}
	\vspace{-0.4cm}
	\end{figure}

	The Parsing generator is an Unet-like network that is responsible to generate a parsing map of the target pose. Specifically, in the training phase, we require a pair of source-target images $ I_S, I_T$, then obtaining human keypoints $ K_S, K_T $ and human parsing map $ P_S, P_T$ respectively by using openpose \cite{cao2019openpose} and PGN \cite{gong2018instance}. The source action key point $K_s$, the target action key point $K_t$, and the source parsing $P_s$ are concatenated together as input and then pass through an encoder-decoder network to output the parsing $P_r$. The encoder is composed of 4 layers of downsampling convolutional layers, the corresponding decoder is composed of 4 layers of upsampling convolutions. Since the parsing map of the target action is very important for the final generation of the target image, and will directly affect the effectiveness of the subsequent image generator. It is obvious that if the effect of the parsing map is not good, it is impossible to generate an ideal image. We found that the generation effect of the parsing map of the target pose of the PISE method is not very good during the experiment. Inspired by \cite{mao2021deep}, we use Res FFT-Conv Block instead of ResBlock, as shown in Figure \ref{fig4}, the difference from traditional ResBlock is that a stream based on a channel-wise FFT \cite{brigham1967fast} to attention to the global context in the frequency domain. It can be seen as a Discrete Fourier Transform (DFT), the 1D version can be expressed as:
	\begin{equation}
		X[k] = \sum_{j=0}^{N-1}x[j]e^{-i \frac{2\pi}{N}kj}
	\end{equation}\label{eqution1}
	Where $X[k]$ represents the spectrum at the frequency $w_k=2\pi k/N$, $i$ is the imaginary unit, and $x[j]$ is the sequence of $N$ complex numbers. It can be seen from the above formula that when n takes any value, the corresponding spectrum has global information, which makes such a block enjoys benefits from modeling both high-frequency and low-frequency discrepancies. Note that an important property of DFT is that $x[j]$ is conjugate symmetric. That is:
	\begin{equation}
		X[N-k] = \sum_{j=0}^{N-1}x[j]e^{-i \frac{2\pi}{N}(N-k)j} = X^*[k]
	\end{equation}
	The symbols here have the same meaning as the symbols in Equation (1). The same can be applied to 2D DTF, namely $X[M-u, N-v] = X^*[u,v]$. In our code, we use the more simplified and efficient FFT in PyTorch instead of DFT. The specific FFT process is as follows:
	\begin{enumerate}
		\item we perform FFT on the real part to obtain the frequency domain feature F(x), where x $\in \mathbb{R}^{H*W*C} $. H, W, and C represent the height, width, and channel of the feature tensor.
		\item the feature extraction operation is performed on the feature F(x) to get feature z through two 1×1 convolutional layers and an activation layer ReLU \cite{li2017convergence} in the middle, where the 1×1 convolution kernel is used for not changing the size of F(x) and getting a wider receptive field. 
		\item applies inverse 2D real FFT to convert back to spatial (time-domain) features.
	\end{enumerate}
	It is worth noting that to make the size of the input and output consistent, we use a width of W/2 when performing the FFT operation.

	\subsection{Image Generator}

 The goal of the image generator is to migrate the textures of human regions of the source pose image to the parsing map of the target pose. The generator used here is similar to PISE, that is, the encoder is first used to extract the regional style texture corresponding to the image $I_s$ of the source pose and the parsing map $P_s$, then use the normalization technique to fit the texture and pose. Since there are invisible areas between the source pose image and the target pose image, joint local and global per-region average pooling is used to extract the regional style features corresponding to the source pose image and the parsing map to fill the invisible area. The joint local and global per-region average pooling are formulated as:
	\begin{equation}
		P(S_{s_j}) = \begin{cases}
			\mathop{avg}\limits_{w,h} (F_i \cdot S_{s_j}), & \sum S_{s_j} > 0  \\
			\mathop{avg}\limits_{w,h} (F_i),& \sum S_{s_j} \leq 0 \\
		\end{cases},
	\end{equation}
	Where $F_i$ is the feature map of Per-region Normalization, $avg(\cdot)$ represents the spatial average pooling, $S_{s_j}$ denotes the semantic map of the source image. To focus on more spatial information, we use Spatial-aware normalization to preserve the spatial information of the source image. Besides that, Resblock here is also replaced with Res FFT-Conv Block to cover more information. Finally, go through a decoder to obtain the desired target pose image $I_T$.
	\subsection{Discriminator}
	In order to distinguish high-resolution real images from generated images, the discriminator needs a large receptive field, which requires a large convolution kernel or a deep convolution layer. Inspired by \cite{wang2018high}, We use a multi-scale residual discriminator, which can harvest a large receptive field with a small amount of network capacity. Its main architecture is VGG-19 \cite{Simonyan2015VeryDC}, and the residual module is used to correspond to the parsing generator. In addition, we deploy spectral normalization in the discriminator to overcome the problem of model instability. Spectral normalization is to decompose the parameter $W$ of each layer of the neural network by SVD \cite{andrews1976singular} and then limit its maximum singular value to 1. Specifically, firstly use SVD to calculate the largest singular value $\sigma(W)$ of $W$, and then divide $W$ by its largest singular value to get the normalized result. The formulas are as follows:
	\begin{equation}
		\sigma(W)=\mathop{max}\limits_{h:h\not=0}\frac{\left \| Wh \right\|_2}{\left \| h \right\|_2}
		\vspace{-0.2cm}
	\end{equation}
	
	\begin{equation}
		W_{SN}=\frac{W}{\sigma(W)}
		\vspace{0cm}
	\end{equation}

	where $\sigma(W)$ is the largest singular value, which is estimated by the power iteration method for simplicity of calculation. In this way, the maximum stretch factor of each layer for the input x will not exceed 1, which makes the discriminator satisfy the Lipschitz continuity \cite{gouk2021regularisation}.
	\subsection{Loss Functions}
	Since there are two generators, and each generator plays a different role, we first train the two generators separately and then perform end-to-end fine-tuning training. 

	\subsubsection{parsing generator loss}For the training of the parsing generator, it can be regarded as supervised learning, and its loss can be written as:
	\begin{equation}
		L_{parsing} = \lambda_{p\ell} L_{\ell_1} + L_{CE}.
		\vspace{-0.2cm}
	\end{equation}
	Where $\lambda_{p\ell}$ is the coefficient of the $\ell_1$ item, and L1 loss can be formulated as:
	\begin{equation}
		L_{l_1} = ||P_g - P_t||_1.
		\vspace{-0.2cm}
	\end{equation}
	$L_{CE}$ is the cross-entropy loss function, which is responsible for generating the correct semantic label (8 in this paper), its expression is:
	\begin{equation}
		L_{CE} = - \frac{1}{N} \sum_{i=0}^{N-1}  P_{t_i}  \log (\textrm{Softmax}(P_{g_i}))
		\vspace{-0.2cm}
	\end{equation}
	Where the $P_{t_i}$ and $P_{g_i}$ represent respectively the parsing map of target and source images. The L1 loss is to generate the correct image at the pixel level. The combined effect of the above two losses makes the parsing generator capable of generating an ideal parsing map.
	\subsubsection{image generator loss}
	In brief, the image generator's job is to texture the generated parsing map, so two main points are involved: the first is that the generated features and the features of the target image are in the same domain. We use a correspondence loss to constrain the generated image features aligned with the target features of the pre-trained VGG-19 in the same domain, and the formula is as follows:
	\begin{equation}
		L_{cor} = ||F_n- {\phi_i}(I_t)||_2.
		\vspace{0cm}
	\end{equation} 
	Where $F_n$ denotes the generated features, $\phi_i(I_t)$ represents the features of the target image from VGG-19. The second is to generate a target image that is as realistic as possible. In our experiments, we used four losses: 
	\begin{equation}
		L_{\ell_1} = ||I_g - I_t||_1.
		\vspace{0.1cm}
	\end{equation}

	\begin{equation}
		L_{perc} = \sum_{i} ||\phi_i(I_t)-\phi_i(I_g)||_1.
		\vspace{-0.1cm}
	\end{equation}
	\begin{equation}
		L_{style} = \sum_{j} ||G^\phi_j(I_t)-G^\phi_j(I_g)||_1.
		\vspace{-0.1cm}
	\end{equation}
	\begin{equation}
		\mathcal{W}[p,q]=\mathop{inf}\limits_{\gamma\in\prod[p,q]}\iint\gamma(x,y)d(x,y)dxdy  \label{wo}
	\end{equation}
	Where $L_{\ell_1}$, $L_{per}$, $L_{style}$, $\mathcal{W}[p,q]$ represent L1 loss, perceptual loss, style loss, Wasserstein distance respectively. $\phi_i(*)$ denotes the feature of $[ReLUi\_1]$ of the pretrained VGG-19 model. $G^\phi_j$ represents the feature of $[ReLU2\_2, ReLU3\_4, ReLU4\_4, ReLU5\_2]$ of the generator. The $ L_{l_1}$ is mainly to align the generated image with the target image at pixel level.  The $L_{perc} $ is to generate more ideal quality images at the human perceptual level. The $L_{style} $ measures the statistical difference between the image generated after the activation layer and the target image. The Wasserstein distance \cite{yin2021unpaired} is to solve the problem that GAN is difficult to train and difficult to converge. The advantage of using the $	\mathcal{W}[p,q] $ to measure the difference between the generated distribution $p$ and the real distribution $q$ is that when there is no intersection or a small intersection between $p$ and $q$. Even Wasserstein distance is not a constant, it can still measure the difference between the two distributions difference, which can alleviate the problem of model collapse. For ease of implementation, we use the Sinkhorn \cite{cuturi2013sinkhorn} distance in PyTorch to replace the Wasserstein distance in our experiments. To sum up, the loss of the image generator can be expressed as:
	\begin{equation}
		\mathcal{L}_{image} = \lambda_c L_{cor} +\lambda_\ell L_{\ell_1} + \lambda_p L_{perc} + \lambda_s \L_{style} + \lambda_a \mathcal{W}
		\vspace{0cm}
	\end{equation}
	where $\lambda_c$, $\lambda_\ell$, $\lambda_p$, $\lambda_s$ and $\lambda_a$ are weights that balance contributions of individual loss terms.
	
	\section{Experiment Setup}
	\subsection{Dataset}
	We conduct experiments on DeepFashion \cite{liu2016deepfashion} dataset, which contains 800000 images, including many male and female models in different poses, various clothes, and buyer shows. We chose DeepFashion In shop Clothes Retrieval Benchmark according to the task of pose transfer, which contains 52712 model and clothes images with a resolution of 256 $\times$ 176. In our experiments, the dataset is split into 110416 pairs by the processing method in PATN \cite{zhu2019progressive}, of which 101966 pairs were used for training. In order to reflect the practicality of our model, the characters in our test set are different from those in the train set.

	\subsection{Implementation Details}
	We use the training strategy used in PISE unless specified. The batch size (16), total training iteration (500000), optimizer (Adam), and initial learning rate (1$\times$$e-4$) are the network training hyperparameters (and the default settings we use). Using the cosine annealing approach, the learning rate is gradually reduced to 1$\times$$e-6$. And we use 4 Tesla-V100 with 16G memory to experiment.

	\subsection{Metrics}
	To assess the effect of our generated images, we employ two typical metrics: First, Learned Perceptual Image Patch Similarity (LPIPS) \cite{zhang2018unreasonable} that is more in line with human perception is employed to measure the perceptional distance between the generated images and the ground-truth images. Second, we compute the error of each pixel between the ground-truth and generated images using the peak signal-to-noise ratio (PSNR).
	
	\section{Experimental Results}

	\subsection{Qualitative Comparison}
	At present, the methods for synthesizing human images are as follows: PATN \cite{zhu2019progressive}, XingGAN \cite{tang2020xinggan}, BiGragh \cite{tang2020bipartite}, ADGAN \cite{men2020controllable}, GFLA \cite{ren2020deep}, PINet \cite{zhang2020human}, and PISE, and the best methods is PISE. We reproduced the results of PISE according to the open-source code provided by its author. The comparisons between the images of our method and the images in PISE are shown in the figure \ref{fig:fig5}. It is worth noting that some of the results are better than those in the PISE paper. That's because we retrained and retested the model with the size 256 $\times$ 176 of input using the author's pretrained model. It can be seen that the results of PISE are better than other methods in some aspects. However, the PISE method fails to generate ideal character details, such as color and clothes. Our method uses Res FFT-Conv Block to retain the global information and hence generate the more realistic color and expression of human images. Specifically, In the first row, the hand obtained by our method is more complete. In the second row, the clothes obtained by our method are more realistic. In the third row, the clothes and hair of our method are more reasonable. In the fourth and fifth rows, our method is more accurate in getting the character's pants color and clothing length. And we use the red box to outline where our method is better than the PISE method.
	\begin{table*}[!t]
	\vspace{-0.2cm}
	\renewcommand{\arraystretch}{1.3}
	\setlength{\tabcolsep}{5mm}
	\begin{center}
		\caption{The effect of Spectral normalization and Wasserstein distance. w/o both: neither spectral normalization nor Wasserstein distance, w/o spectral norm: with Wasserstein distance bot without spectral normalization, w/o Wasserstein distance: with spectral normalization but without Wasserstein distance, Full: with Wasserstein distance and spectral normalization.}\label{tab:tab4}
		\begin{tabular}{|c|c|c|c|c|}
			\hline
			\diagbox[width=3cm, height=0.7cm]{metrics}{model}  & w/o both & w/o spectral norm& w/o Wasserstein distance  & Full \\
			\hline
			iterations of training &  362000 & 294000 & 268000 &  253000 \\
			\hline
			LPIPS $\downarrow$ & 0.1668 & 0.1664 & 0.1656  & 0.1644 \\
			\hline
			PSNR $\uparrow$ &  31.38 &   31.40 & 31.39  & 31.40\\
			\hline
		\end{tabular}
	\end{center}
	\vspace{-0.2cm}
	\end{table*}
	\subsection{Quantitative Comparison}
\begin{figure}[H]
	\centering
	\includegraphics[width=0.8\linewidth]{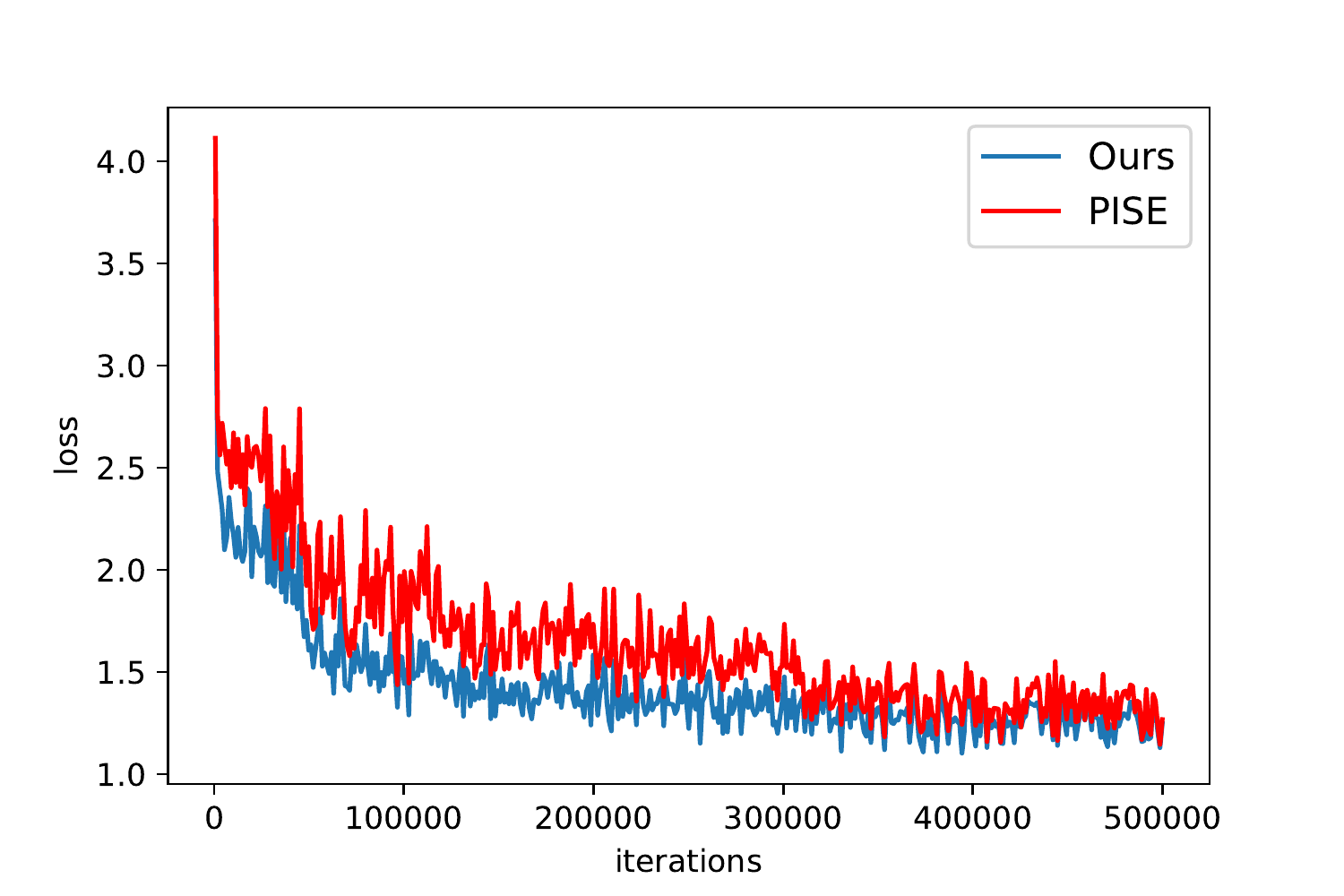}
	\caption{Training loss of PISE and ours. Our method not only makes the training more stable but also reduces the loss.}
	\label{fig:fig6}
	\vspace{-0.2cm}
\end{figure}

	As shown in the TABLE \ref{tab:tab2}, we use the images generated in the test set and the real images to calculate the LPIPS, PSNR. The results show that our LPIPS metric is 0.0415 lower than the best method, indicating that our method focuses on more information. This also reflects that the Res FFT-Conv Block can obtain rich high-frequency and low-frequency information. In addition, the PSNR score, which reflects the image signal-to-noise ratio, is also improved, reflecting that our model not only generates more realistic images, but also better aligns the shape and texture of the images. Since we apply Wasserstein distance and spectral normalization to make GAN training easier and faster to converge, our training convergence iterations and training loss are lower than PISE in Figure \ref{fig:fig6}, indicating that our method can alleviate GAN training difficulty.

	\vspace{-0.2cm}
	\begin{table}[h]
		\renewcommand{\arraystretch}{1.3}
		\small%
		\setlength{\tabcolsep}{5.5mm}
		\begin{center}
			\caption{Quantitative comparison with state-of-the-art methods. Our method gets the best score in terms of both LPIPS and PSNR.}\label{tab:tab2}
			\begin{tabular}{|c|c|c|}
				\hline
				{Model} &  LPIPS $\downarrow$ & PSNR $\uparrow$ \\
				\hline
				PATN \cite{zhu2019progressive}  & 0.2520 & 31.16 \\
				BiGraph \cite{tang2020bipartite}  & 0.2428 & 31.38 \\
				XingGAN \cite{tang2020xinggan} & 0.2914 & 31.08 \\
				GFLA \cite{ren2020deep}  & 0.2219 & 31.28 \\
				ADGAN \cite{men2020controllable}  & 0.2242 & 31.30\\
				PINet \cite{zhang2020human}  & 0.2152 & 31.31\\
				PISE \cite{zhang2021pise}  & 0.1844 & 31.38  \\
				\hline
				Ours   & \textbf{0.1644} & \textbf{31.40} \\
				\hline
			\end{tabular}
			
		\end{center}
		\vspace{-0.4cm}
	\end{table}

	\subsection{Human image pose transfer} 
\begin{figure}[H]
		\centering
		\includegraphics[width=0.8\linewidth]{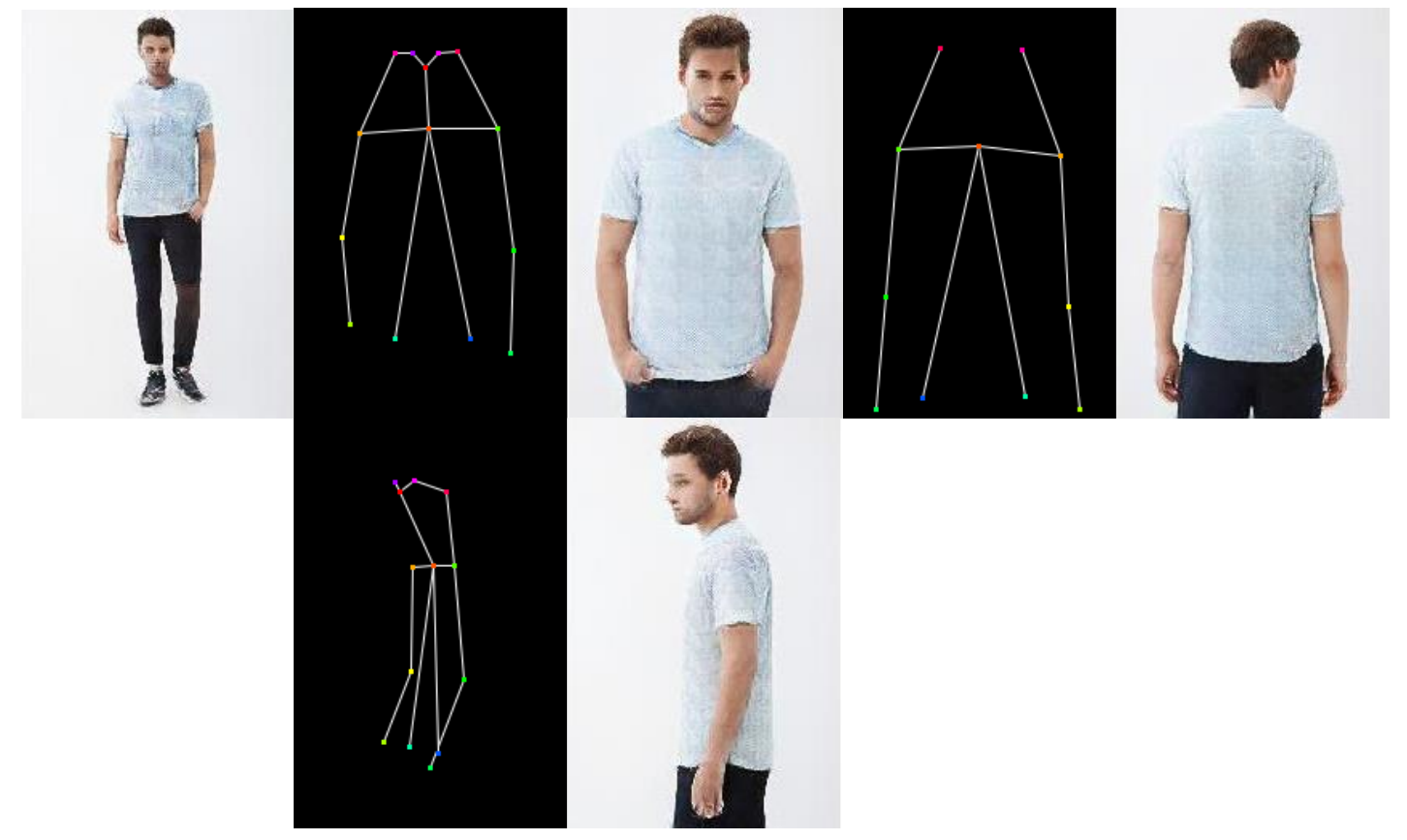}
		\caption{Our results of human image synthesis in different poses.}
		\label{fig7}
		\vspace{0cm}
	\end{figure}

\begin{figure*}
		\centering
		\includegraphics[width=1.0\linewidth]{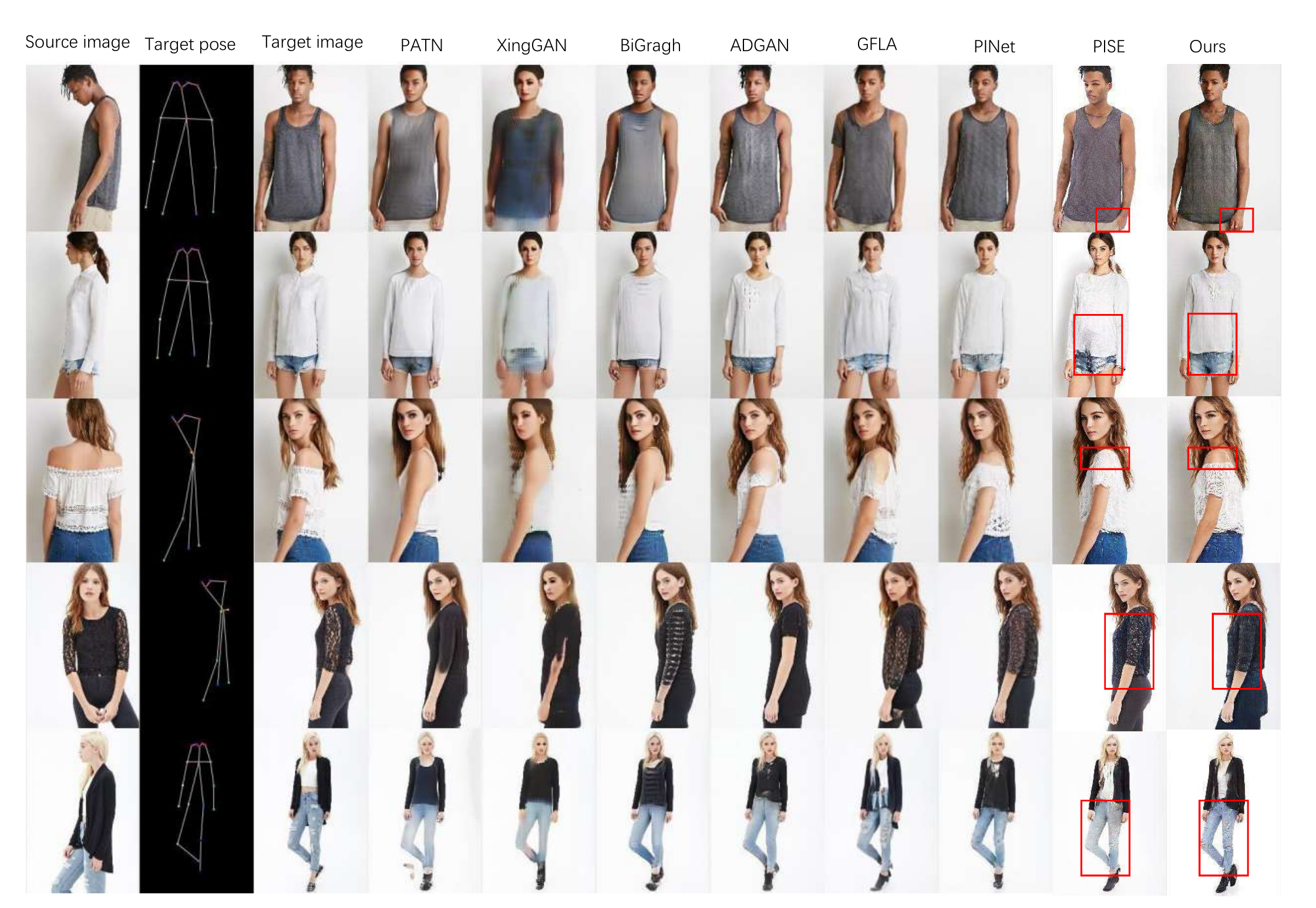}	
			\caption{Overall, our method is the best. Specifically, our method has the brightest colors and the most realistic expressions and is the closest to ground-truth. As you can see from the red box that is outlined in the figure. }
		\label{fig:fig5}
	\end{figure*}

	Human image pose transfer means that given a source image, a target image with different poses is then generated based on different poses from the test images. As shown in Figure \ref{fig7}, our model can generate realistic images with different poses and fine details.
	\section{Ablation Study}
	
	To explore the source of the improved model performance of our method, we did an extensive ablation study to analyze the effect of each method on our experiments. Specifically, we analyze the effect of Res FFT-Conv Block and Wasserstein distance and spectral normalization separately. For clarity, we use quantitative metrics to measure, and we use the number of iterations to measure the convergence speed in the experiments.
	\subsubsection{The effect of spectral normalization and Wasserstein distance}
	We tested the effects of spectral normalization and Wasserstein distance on the experiments respectively. Following the principle of control variables, we only changed spectral normalization and Wasserstein distance in the model without changing the Res FFT-Conv Block. The experimental results are shown in TABLE \ref{tab:tab4}. Both spectral normalization and Wasserstein distance can reduce the number of iterations for training to a certain extent (68000 for spectral norm, 94000 for Wasserstein distance), which means that they can improve the convergence speed of the model. Because Wasserstein distance restricts the model convergence at the level of the loss function, the optimization effect is more obvious. The result of Full (with spectral normalization and Wasserstein distance) shows that the two stacking effects are better. We argue that the two tricks belong to different levels of optimization, which makes better in the stacking effect.
	\vspace{0.2cm}
	\subsubsection{The Effect of Res FFT-Conv Block}
	 To explore the role of Res FFT-Conv Block, we compare the model with and without Res FFT-Conv Block, noting that other conditions (spectral normalization and Wasserstein distance) are the same. The results are shown in table \ref{table3}. The LPIPS index with Res FFT-Conv Block is much better, and the number of iterations of training is substantially unchanged, which shows that Res FFT-Conv Block can significantly improve the quality of generation, but does not help much for stable training.
	\begin{table}[!h]
		\renewcommand{\arraystretch}{1.3}
		\setlength{\tabcolsep}{2.2mm}
		\begin{center}
			\caption{The effect of Res FFT-Conv Block. w/o: without, w: with}
			\begin{tabular}{|c|c|c|}
				\hline
				\diagbox[width=3cm, height=0.7cm]{metrics}{model} & w/o Res FFT-Conv& w Res FFT-Conv \\
				\hline
				iterations of training & 256000 & 253000  \\
				\hline
				LPIPS $\downarrow$ & 0.1921 & 0.1644  \\
				\hline
				PSNR $ \uparrow$ &  31.39 &   31.40 \\
				\hline
			\end{tabular}
			\label{table3}
		\end{center}
		\vspace{-0.2cm}
	\end{table}
	
	\section{Conclusion}
	We find the shortcomings of this method by analyzing the most recently developed human image synthesis model PISE, and then improve it. Specifically, we improve the quality of generated images by introducing Res FFT-Conv Block instead of ordinary ResBlock. For improving the convergence speed of training, we employ Wasserstein distance instead of traditional JS divergence and use spectral normalization to improve the ability of the discriminator.  Experiments show that our method achieves the best effect both in quality and quantity.
	\section*{Acknowledgement}
	This paper is supported by the Key Research and Development Program of Guangdong Province under grant No.
	2021B0101400003. Corresponding author is Shijing Si from Ping An Technology (Shenzhen) Co., Ltd (shijing.si@outlook.com).

	\bibliographystyle{IEEEtran}
	\footnotesize
	\balance
	\bibliography{refs}
	
\end{document}